\documentclass[10pt,twocolumn,letterpaper]{article}

\usepackage{3dv}
\usepackage{times}
\usepackage{epsfig}
\usepackage{graphicx}
\usepackage{amsmath}
\usepackage{amssymb}
\usepackage{gensymb}
\usepackage{tabularx}
\usepackage{xcolor}
\usepackage{subcaption}
\usepackage{hyperref}
\usepackage{makecell}
\usepackage{mathptmx}
\usepackage{balance}

\DeclareMathOperator{\atantwo}{atan2}

\definecolor{res1color}{rgb}{0, 0.670, 0.184}
\definecolor{res2color}{rgb}{1, 0.568, 0.188}

\threedvfinalcopy 

\def\threedvPaperID{19} 

\ifthreedvfinal\fi
\begin{document}

\title{The Double Sphere Camera Model}

\author{Vladyslav Usenko, Nikolaus Demmel and Daniel Cremers\\
Technical University of Munich\\
{\tt\small \{usenko, demmeln, cremers\}@in.tum.de}
}



\maketitle

\begin{abstract}
  Vision-based motion estimation and 3D reconstruction, which have numerous applications (e.g., autonomous driving, navigation systems for airborne devices and augmented reality) are receiving significant research attention. To increase the accuracy and robustness, several researchers have recently demonstrated the benefit of using large field-of-view cameras for such applications.
  
  In this paper, we provide an extensive review of existing models for large field-of-view cameras. For each model we provide projection and unprojection functions and the subspace of points that result in valid projection. Then, we propose the Double Sphere camera model that well fits with large field-of-view lenses, is computationally inexpensive and has a closed-form inverse. We evaluate the model using a calibration dataset with several different lenses and compare the models using the metrics that are relevant for Visual Odometry, i.e., reprojection error, as well as computation time for projection and unprojection functions and their Jacobians. We also provide qualitative results and discuss the performance of all models.
  
\end{abstract}


\section{Introduction}

Visual Odometry and Simultaneous Localization and Mapping are becoming important for numerous applications. To increase the accuracy and robustness of these methods, both hardware and software must be improved.

Several issues can be addressed by the use of large field-of-view cameras. First, with a large field-of-view, it is easier to capture more textured regions in the environment, which is required for stable vision-based motion estimation. Second, with a large field-of-view, large camera motions can be mapped to smaller pixel motions compared to cameras with a smaller field-of-view at the same resolution. This ensures small optical flow between consecutive frames, which is particularly beneficial for direct methods.

\begin{figure}[t]
\begin{center}
\def\svgwidth{3in}
\resizebox{0.9\linewidth}{!}{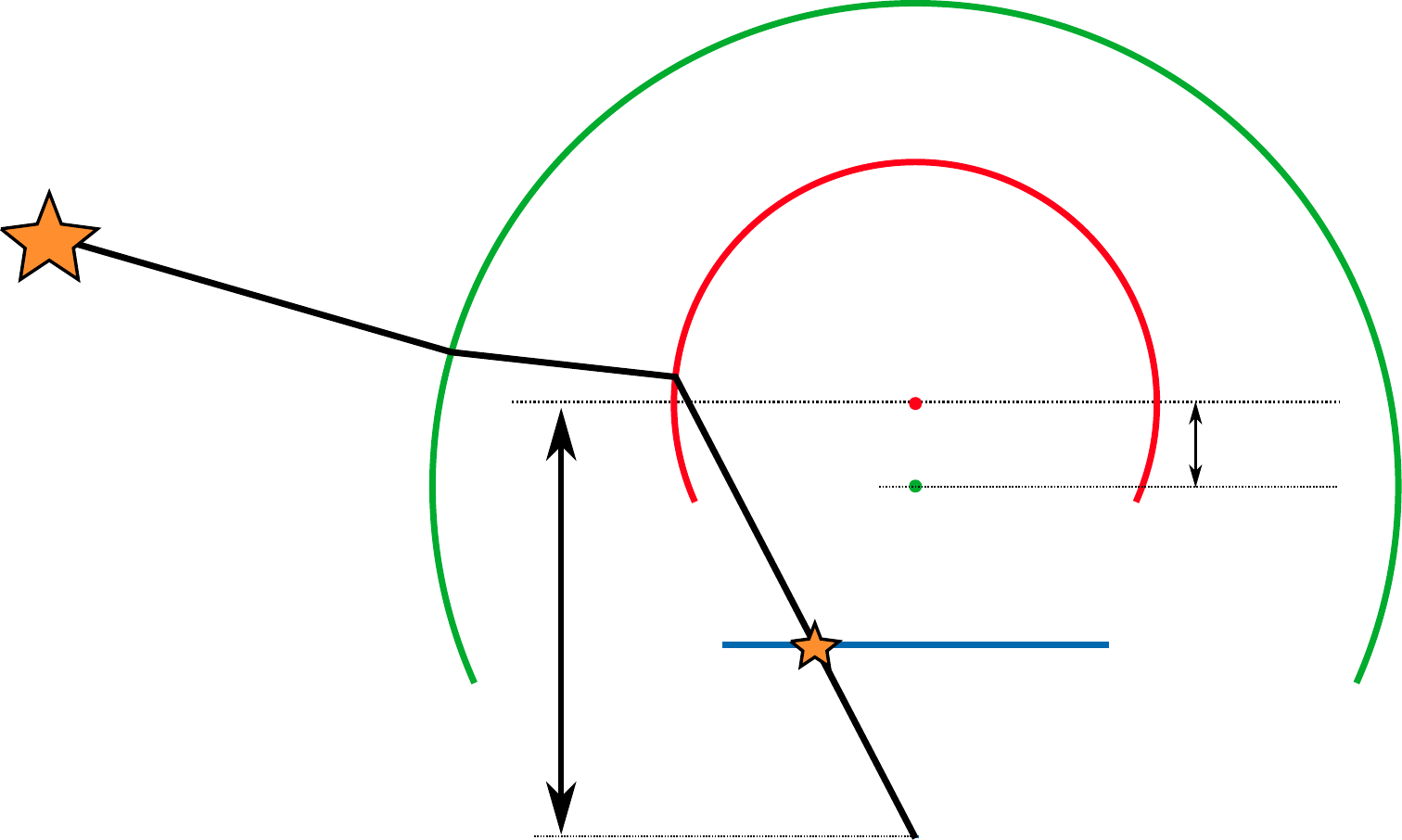}
\end{center}
   \caption{The proposed Double Sphere (\textbf{DS}) projection model. Initially, the point is projected onto the first sphere (green) and then onto the second sphere, which is shifted with respect to the first sphere by $\xi$ (red). Then, the point is projected onto the image plane of a pinhole camera that is shifted by $\frac{\alpha}{1-\alpha}$ from the second sphere. The image below is the reprojection of the pattern corners after calibration using the proposed DS model, which indicates that the model fits the lens well.}
\label{fig:double_sphere}
\vspace{8pt}
\includegraphics[width=1.0\linewidth]{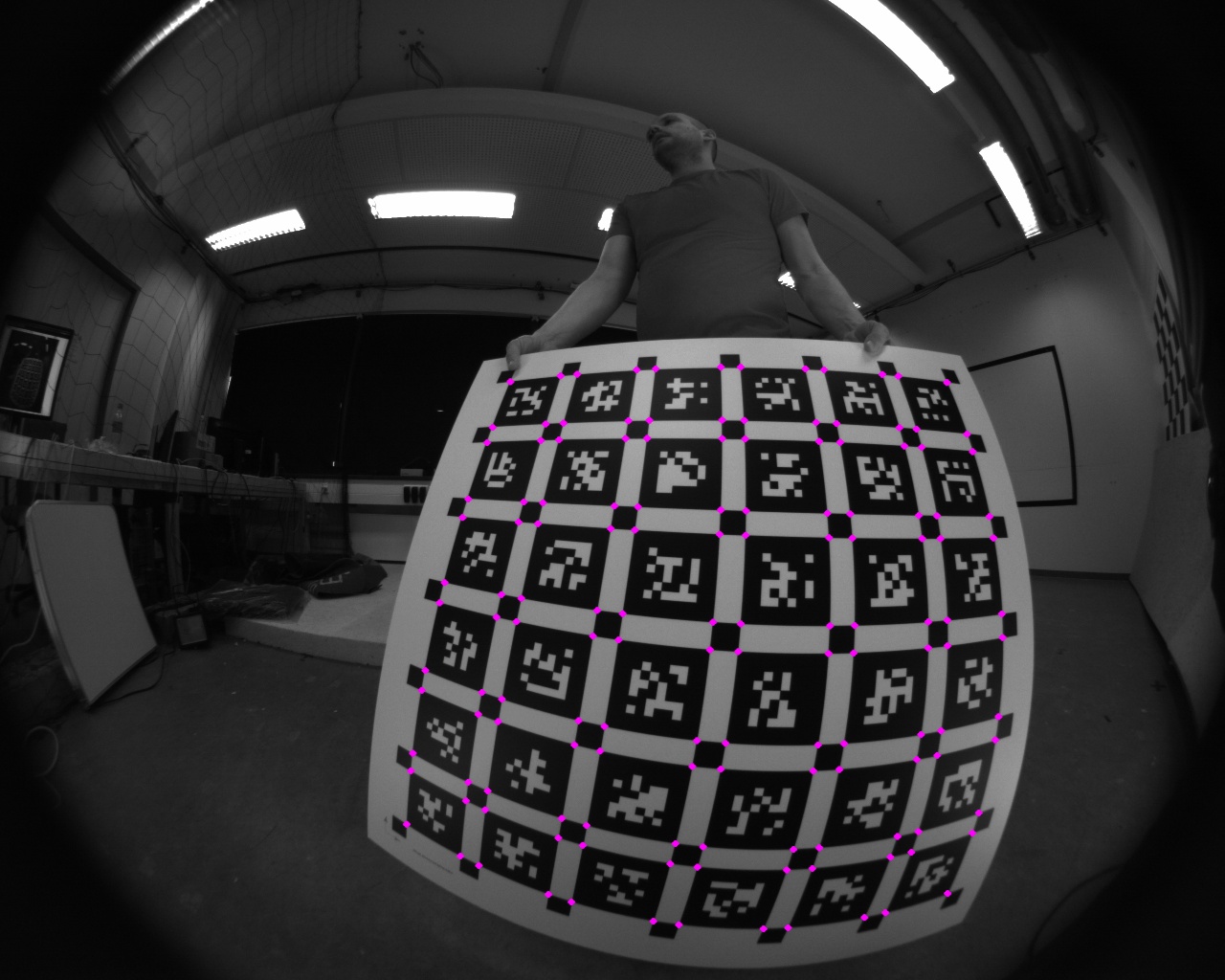}
\end{figure}

Previous studies have demonstrated that a large field-of-view is beneficial for vision-based motion estimation \cite{zhang16} \cite{rituerto2010comparison}. Catadioptric cameras are mechanically complex and expensive; however fisheye lenses are small, lightweight, and widely available on the consumer market. Thus, in this paper we focus on fisheye lenses and models that describe their projection.

The reminder of this paper is organized as follows. In Section \ref{sec:related_work} we provide an extensive review of camera models that can be used with fisheye lenses. To make the paper self-contained we provide the projection and unprojection functions and define the subspace of valid projections for each model. In Section \ref{sec:double_sphere}, we propose a novel projection model for fisheye cameras that has the following advantages.

\begin{itemize}
    \item The proposed projection model is well suited to represent the distortions of fisheye lenses.
    \item The proposed model does not require computationally expensive trigonometric operations for projection and unprojection.
    \item Differing from projection models based on higher order polynomials \cite{kannala2006generic} \cite{scaramuzza2006}, that use iterative methods to unproject points, the inverse of the projection function exists in a closed form. 
\end{itemize}

In Section \ref{sec:evaluation}, we evaluate all presented models with respect to metrics that are relevant for vision-based motion estimation. Here, we use a dataset collected using several different lenses to evaluate the reprojection error for each model. We also present the computation time required for projection and unprojection functions and the time required to compute Jacobians relative to their arguments.

The datasets used in this study together with the open-source implementation of the proposed model are available  on the project page:

\begin{center}
\color{purple}
\urlstyle{tt}
\textbf{\small
        \url{https://vision.in.tum.de/research/vslam/double-sphere}}
\end{center}
\vspace{-0.3cm}

\section{Related Work}
\label{sec:related_work}

We define the notations used in this paper prior to reviewing existing camera models that can be used with fisheye lenses.
We use lowercase letters to denote scalars, e.g., $u$, bold uppercase letters to denote matrices, e.g., $\bf R$, and bold lowercase letters for vectors, e.g., $\bf x$.

Generally, we represent pixel coordinates as $\mathbf{u} = \left[ u, v \right]^T \in \Theta \subset \mathbb{R}^2$, where $\Theta$ denotes the image domain to which points can be projected to.
3D point coordinates are denoted $\mathbf{x} = \left[ x, y, z \right]^T \in \Omega \subset \mathbb{R}^3$, where $\Omega$ denotes the set of 3D points that result in valid projections.

For all camera models we assume all projections cross a single point (i.e., central projection) that defines the position of the camera coordinate frame. The orientation of the camera frame is defined as follows. The $z$ axis is aligned with the principal axis of the camera, and two other orthogonal directions $(x,y)$ align with the corresponding axes of the image plane. We define a coordinate frame rigidly attached to the calibration pattern such that the transformation ~${\bf T}_{ca_n} \in SE(3)$, which is a matrix of the special Euclidean group, transforms a 3D coordinate from the calibration pattern coordinate system to the camera coordinate system for image $n$.

Generally, a camera projection function is a mapping $\pi : \Omega \to \Theta$.
Its inverse $\pi^{-1} : \Theta \to \mathbb{S}^2$ unprojects image coordinates to the bearing vector of unit length, which defines a ray by which all points are projected to these image coordinates.

For all camera models discussed in this section, we provide definitions of $\pi$, $\pi^{-1}$, the vector of intrinsic parameters $\mathbf{i}$, $\Omega$ and $\Theta$. 

\begin{figure}
\centering{
\def\svgwidth{3in}
\resizebox{0.9\linewidth}{!}{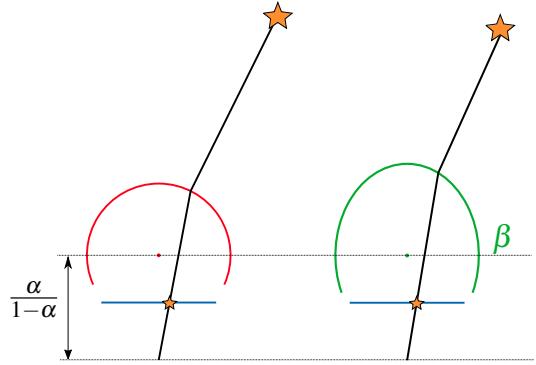}
\caption{Schematic representation of the Unified Camera Model (\textbf{UCM}) and Extended Unified Camera Model (\textbf{EUCM}). First a 3D point is projected onto a unit sphere and then projected onto the image plane of the pinhole camera shifted by $\frac{\alpha}{1-\alpha}$ from the center of the sphere. In the EUCM, the sphere is transformed to an ellipsoid using the coefficient $\beta$. }
\label{fig:ucm_eucm}
}
\end{figure}

\subsection{Pinhole Camera Model}
The pinhole camera model has four parameters $\mathbf{i} = \left[f_x, f_y, c_x, c_y\right]^T$ with a projection function that is defined as follows:

\begin{eqnarray}
\pi(\mathbf{x}, \mathbf{i})=
\begin{bmatrix}
    f_x\frac{x}{z} \\ 
    f_y\frac{y}{z} \\
\end{bmatrix}
+
\begin{bmatrix}
c_x \\ 
c_y\\
\end{bmatrix},
\end{eqnarray}

It is easy to see that projection is defined for $\Omega = \{\mathbf{x} \in \mathbb{R}^3 ~|~ z > 0\}$, which theoretically limits the field-of-view to less than 180\degree. However, in practice, even when distortion model is added the pinhole camera demonstrates suboptimal performance for a field-of-view greater than 120\degree.

We can use the following function to unproject a point:
\begin{align}
    \pi^{-1}(\mathbf{u}, \mathbf{i}) &= 
\frac{1}{\sqrt{m_x^2 + m_y^2 + 1}}
\begin{bmatrix}
m_x\\ 
m_y\\
1\\
\end{bmatrix}\\
    m_x &= \frac{u -  c_x}{f_x}, \\
    m_y &= \frac{v -  c_y}{f_y},
\end{align}
where unprojection is defined for $\Theta = \mathbb{R}^2$.

\subsection{Unified Camera Model}

The unified camera model (\textbf{UCM}) has five parameters $\mathbf{i} = \left[\gamma_x, \gamma_y, c_x, c_y, \xi\right]^T$ and is typically used with catadioptric cameras \cite{mei}.
A previous study \cite{geyer} has shown that the UCM can represent systems with parabolic, hyperbolic, elliptic and planar mirrors. This model can also be applied to cameras with fisheye lenses \cite{ying}. However, it does not fit most fisheye lenses perfectly; thus, an additional distortion model is often added.

In the UCM, projection is defined as follows:
\begin{align}
    \pi(\mathbf{x}, \mathbf{i}) &=
    \begin{bmatrix}
    \gamma_x{\frac{x}{\xi d + z}}\\
    \gamma_y{\frac{y}{\xi d + z}}\\
    \end{bmatrix}
    +
    \begin{bmatrix}
    c_x \\ 
    c_y\\
    \end{bmatrix},\\
    d &= \sqrt{x^2 + y^2 + z^2}.
    \label{eq:ucm_mei}
\end{align}
In this model, a point is first projected onto the unit sphere and then onto the image plane of the pinhole camera, which is shifted by $\xi$ from the center of the unit sphere.

For practical applications we propose to rewrite this model as follows:
\begin{align}
    \pi(\mathbf{x}, \mathbf{i}) &=
    \begin{bmatrix}
    f_x{\frac{x}{\alpha d + (1-\alpha) z}}\\
    f_y{\frac{y}{\alpha d + (1-\alpha) z}}\\
    \end{bmatrix}
    +
    \begin{bmatrix}
    c_x \\ 
    c_y\\
    \end{bmatrix}.
    \label{eq:ucm_ours}
\end{align}
This formulation of the model also has five parameters $\mathbf{i} = \left[f_x, f_y, c_x, c_y, \alpha\right]^T$, $\alpha~\in~[0,1]$ and is mathematically equivalent to the previous one ($\xi = \frac{\alpha}{1-\alpha}, \gamma_x = \frac{f_x}{1-\alpha}, \gamma_y = \frac{f_y}{1-\alpha}$). However, as discussed in Section \ref{sec:comparison_ucm}, it has better numerical properties. Note that for $\alpha = 0$, the model degrades to the pinhole model.

The set of 3D points that result in valid projections is defined as follows:
\begin{align}
\Omega &= \{\mathbf{x} \in \mathbb{R}^3 ~|~ z > -wd \}, \\
w &= \begin{cases} \frac{\alpha}{1-\alpha}, & \mbox{if } \alpha \le 0.5, \\  \frac{1-\alpha}{\alpha} & \mbox{if } \alpha > 0.5, \end{cases}
\end{align}
where (for $\alpha > 0.5$) $w$ represents the sine of the angle between the horizontal axis on schematic plot (Figure \ref{fig:ucm_eucm}) and the perpendicular to the tangent of the circle from the focal point of the pinhole camera.

The unprojection function is defined as follows:
\begin{align}
    \pi^{-1}(\mathbf{u}, \mathbf{i}) &= 
    \frac{\xi + \sqrt{1 + (1 - \xi^2) r^2}} {1 + r^2}
    \begin{bmatrix}
    m_x \\ 
    m_y \\
    1 \\
    \end{bmatrix} -
    \begin{bmatrix}
    0 \\ 
    0 \\
    \xi \\
    \end{bmatrix}, \\
    m_x &= \frac{u - c_x}{f_x}(1-\alpha), \\
    m_y &= \frac{v - c_y}{f_y}(1-\alpha), \\
    r^2 &= m_x^2 + m_y^2, \\
    \xi &= \frac{\alpha}{1-\alpha},
\end{align}
where $\Theta$ is defined as follows.
\begin{align}
\Theta &= 
\begin{cases} 
\mathbb{R}^2 & \mbox{if } \alpha \le 0.5 \\
\{ \mathbf{u} \in \mathbb{R}^2 ~|~ r^2 \le \frac{(1-\alpha)^2}{2\alpha - 1} \}  & \mbox{if } \alpha > 0.5  
\end{cases}
\end{align}

\begin{figure}
\centering{
\def\svgwidth{1in}
\resizebox{0.25\linewidth}{!}{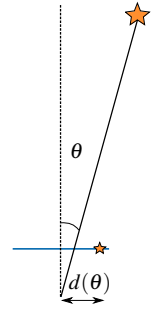}
\caption{Schematic representation of the Kannala-Brandt Camera model (\textbf{KB}). The displacement of the projection from the optical center is proportional to $d(\theta)$, which is a polynomial function of the angle between the point and optical axis $\theta$.  }
\label{fig:kb}
}
\end{figure}

\subsection{Extended Unified Camera Model}

A previous study \cite{khomutenko} extended the unified camera model (\textbf{EUCM}) to have six parameters $\mathbf{i} = \left[f_x, f_y, c_x, c_y, \alpha, \beta \right]^T$, $\alpha~\in~[0,1]$, $\beta~>~0$ and defined the following projection function.
\begin{align}
    \pi(\mathbf{x}, \mathbf{i}) &=
    \begin{bmatrix}
    f_x{\frac{x}{\alpha d + (1-\alpha) z}}\\
    f_y{\frac{y}{\alpha d + (1-\alpha) z}}\\
    \end{bmatrix}
    +
    \begin{bmatrix}
    c_x \\ 
    c_y\\
    \end{bmatrix}, \\
    d &= \sqrt{\beta(x^2 + y^2) + z^2}. \label{eq:d_eucm}
\end{align}

The EUCM can be interpreted as a generalization of the UCM where the point is projected onto an ellipsoid symmetric around the $z$ axis (Figure \ref{fig:ucm_eucm}). That study also indicated that when treating the model as a projection on a quadratic surface followed by orthographic projection on the image plane the model is complete in the sense that it can represent all possible quadratic surfaces.

With EUCM, a set $\Omega$ is defined similar to the UCM, with the difference that $d$ is computed by Eq. \ref{eq:d_eucm}. Note that the EUCM degrades to a regular UCM for $\beta=1$.

As mentioned previously, the EUCM projects on the ellipsoid. Therefore, the unit length vector for unprojection cannot be obtained directly; consequently, we must employ normalization. The unprojection function is defined as follows:

\begin{align}
    \pi^{-1}(\mathbf{u}, \mathbf{i}) &= 
    \frac{1} {\sqrt{m_x^2 + m_y^2 + m_z^2}}
    \begin{bmatrix}
    m_x \\ 
    m_y \\
    m_z \\
    \end{bmatrix}, \\
    m_x &= \frac{u - c_x}{f_x}, \\
    m_y &= \frac{v - c_y}{f_y}, \\
    r^2 &= m_x^2 + m_y^2, \\
    m_z &= \frac{1 - \beta \alpha^2 r^2}{\alpha \sqrt{1 - (2\alpha - 1) \beta r^2} + (1 - \alpha)},
\end{align}
where $\Theta$ is defined as follows.
\begin{align}
\Theta &= \begin{cases}
\mathbb{R}^2 & \mbox{if } \alpha \le 0.5 \\
\{ \mathbf{u} \in \mathbb{R}^2 ~|~ r^2 \le \frac{1}{\beta(2\alpha-1)} \}  & \mbox{if } \alpha > 0.5 
\end{cases}
\end{align}

\subsection{Kannala-Brandt Camera Model}

The previous study \cite{kannala2006generic} proposed the Kannala-Brandt (\textbf{KB}) model, which is a generic camera model that well fits regular, wide angle and fisheye lenses. The KB model assumes that the distance from the optical center of the image to the projected point is proportional to the polynomial of the angle between the point and the principal axis (Figure \ref{fig:kb}). We evaluate two versions of the KB model, i.e.,: with six parameters $\mathbf{i} = \left[f_x, f_y, c_x, c_y, k_1, k_2\right]^T$ and eight parameters $\mathbf{i} = \left[f_x, f_y, c_x, c_y, k_1, k_2, k_3, k_4\right]^T$. The projection function of the KB model is defined as follows:
\begin{align}
    \pi(\mathbf{x}, \mathbf{i}) &=
    \begin{bmatrix}
    f_x ~ d(\theta) ~ \frac{x}{r}\\
    f_y ~ d(\theta) ~ \frac{y}{r}\\
    \end{bmatrix}
    +
    \begin{bmatrix}
    c_x \\ 
    c_y\\
    \end{bmatrix},\\
    r &= \sqrt{x^2 + y^2}, \\
    \theta &= \atantwo(r, z), \\
    d(\theta) &= \theta + k_1 \theta^3 + k_2 \theta^5 + k_3 \theta^7 + k_4 \theta^9,
\end{align}
assuming that polynomial $d(\theta)$ is monotonic $\Omega = \mathbb{R}^3 \setminus [0,0,0]^T$.

The unprojection function of the KB model requires finding the root of a high-order polynomial to recover angle $\theta$ from $d(\theta)$. This can be achieved through an iterative optimization, e.g., Newton's method. The unprojection function can be expressed as follows:

\begin{align}
    \pi^{-1}(\mathbf{u}, \mathbf{i}) &= 
    \begin{bmatrix}
    \sin(\theta^{*}) ~ \frac{m_x}{r_u}  \\ 
    \sin(\theta^{*}) ~ \frac{m_y}{r_u}  \\
    \cos(\theta^{*}) \\
    \end{bmatrix}, \\
    m_x &= \frac{u - c_x}{f_x}, \\
    m_y &= \frac{v - c_y}{f_y}, \\
    r_u &= \sqrt{m_x^2 + m_y^2}, \\
    \theta^{*} &= d^{-1}(r_u),
\end{align}
where $\theta^{*}$ is the solution of $d(\theta)=r_u$. If $d(\theta)$ is monotonic $\Theta = \mathbb{R}^2$.

The KB model is sometimes used as a distortion model for a pinhole camera, e.g., the \emph{equidistant} distortion model in Kalibr\footnote{\url{https://github.com/ethz-asl/kalibr}\label{foot:kalibr}} \cite{furgale2013} or the \emph{fisheye} camera model in OpenCV\footnote{\url{https://github.com/opencv/opencv}}.
The model is mathematically the same; however, since it first projects the point using the pinhole model and then applies distortion, it has a singularity at $z=0$, which makes it unsuitable for fisheye lenses with field-of-view close to 180\degree when implemented is this manner.

Several other models for large field-of-view lenses based on high-order polynomials exist. For example, the main differences between \cite{scaramuzza2006} and the KB model are as follows: the model calibrates two separate polynomials for projection and unprojection to provide a closed-form solution for both, and for projection it uses the angle between the image plane and the point rather than of the angle between the optical axis and the point. We expect this model to have similar performance and do not explicitly include it in our evaluation.

\begin{figure}
\centering{
\def\svgwidth{2in}
\resizebox{0.5\linewidth}{!}{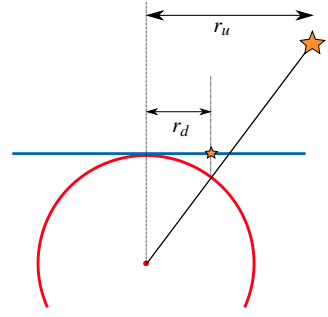}
\caption{Schematic representation of the Field-of-View Camera model (\textbf{FOV}). Displacement of the projection from the optical center is proportional to the angle between the point and optical axis  }
\label{fig:fov}
}
\end{figure}

\subsection{Field-of-View Camera Model}

A previously proposed Field-of-view camera model (\textbf{FOV}) \cite{devernay2001straight}, has five parameters $\mathbf{i} = \left[f_x, f_y, c_x, c_y, w\right]^T$ and assumes the distance between an image point and the principal point is typically approximately proportional to the angle between the corresponding 3D point and the optical axis (Figure \ref{fig:fov}). According to authors, parameter $w$ approximately corresponds to the field-of-view of an ideal fisheye lens. The projection function for this model is defined as follows:

\begin{align}
    \pi(\mathbf{x}, \mathbf{i}) &=
    \begin{bmatrix}
    f_x ~ r_d ~ \frac{x}{r_u}\\
    f_y ~ r_d ~ \frac{y}{r_u}\\
    \end{bmatrix}
    +
    \begin{bmatrix}
    c_x \\ 
    c_y\\
    \end{bmatrix},\\
    r_u &= \sqrt{x^2 + y^2}, \\
    r_d &= \frac{\atantwo(2 r_u \tan{\frac{w}{2}}, z)}{w},
\end{align}
where $\Omega = \mathbb{R}^3 \setminus [0,0,0]^T$.

The FOV model has a closed-form solution for unprojecting the points, which is defined as follows:

\begin{figure}[t]
\begin{center}
\includegraphics[width=\linewidth]{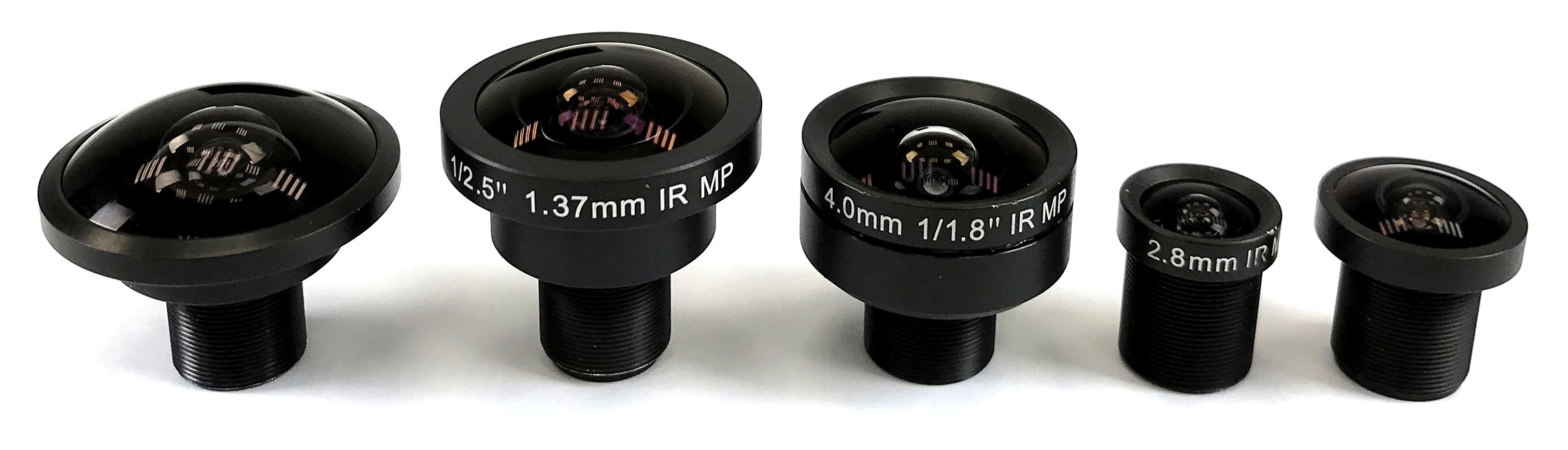}
\end{center}
   \caption{Lenses used to evaluate camera models; left to right: BF2M2020S23 (195\degree), BF5M13720 (183\degree), BM4018S118 (126\degree), BM2820 (122\degree), and GoPro replacement lens (150\degree).}
\label{fig:lenses}
\end{figure}

\begin{align}
    \pi^{-1}(\mathbf{u}, \mathbf{i}) &= 
    \begin{bmatrix}
    m_x \frac{\sin(r_d w)}{ 2 r_d \tan{\frac{w}{2}}}  \\ 
    m_y \frac{\sin(r_d w)}{ 2 r_d \tan{\frac{w}{2}}}  \\
    \cos(r_d w) \\
    \end{bmatrix}, \\
    m_x &= \frac{u - c_x}{f_x}, \\
    m_y &= \frac{v - c_y}{f_y}, \\
    r_d &= \sqrt{m_x^2 + m_y^2},
\end{align}
where $\Theta = \mathbb{R}^2$.

Similar to the KB model, the FOV model can be used as a distortion model for a pinhole camera.

\begin{table*}[t]
\resizebox{\textwidth}{!}{%
\begin{tabular}{ | p{2.5cm} | p{2.2cm} | p{2.4cm} | p{2cm} | p{2cm} | p{2.2cm}  | p{2cm} |}
\hline
\textbf{\makecell{Dataset}} &
\textbf{\makecell{UCM \cite{mei} \\ 5 parameters}} &
\textbf{\makecell{FOV \cite{devernay2001straight} \\ 5 parameters}}  &
\textbf{\makecell{DS (Ours) \\ 6 parameters}} & 
\textbf{\makecell{EUCM \cite{khomutenko} \\ 6 parameters}} &
\textbf{\makecell{KB \cite{kannala2006generic} \\ 6 parameters}} & 
\textbf{\makecell{KB \cite{kannala2006generic} \\ 8 parameters}}  \\
\hline
BF2M2020S23-1 & 0.236 (63.13\%) & 0.417 (187.90\%) & 0.145 (0.35\%) & \textcolor{res2color}{0.145 (0.30\%)}  & 0.164 (13.53\%) & \textcolor{res1color}{0.145 (0.00\%)} \\
BF2M2020S23-2 & 0.250 (59.94\%) & 0.490 (213.34\%) & \textcolor{res2color}{0.157 (0.23\%)}  & 0.157 (0.49\%) & 0.180 (15.43\%) & \textcolor{res1color}{0.156 (0.00\%)} \\
BF2M2020S23-3 & 0.277 (53.99\%) & 0.454 (151.81\%) & \textcolor{res2color}{0.180 (0.11\%)}  & 0.181 (0.47\%) & 0.202 (11.91\%) & \textcolor{res1color}{0.180 (0.00\%)} \\
BF5M13720-1 & 0.228 (49.53\%) & 0.307 (101.14\%) & \textcolor{res1color}{0.153 (0.00\%)}  & 0.154 (0.51\%) & 0.161 (5.41\%) & \textcolor{res2color}{0.153 (0.03\%)} \\
BF5M13720-2 & 0.250 (48.68\%) & 0.379 (124.91\%) & \textcolor{res2color}{0.169 (0.64\%)}  & 0.171 (1.73\%) & 0.183 (8.39\%) & \textcolor{res1color}{0.168 (0.00\%)} \\
BF5M13720-3 & 0.252 (54.99\%) & 0.386 (137.56\%) & \textcolor{res2color}{0.163 (0.56\%)}  & 0.165 (1.64\%) & 0.176 (8.51\%) & \textcolor{res1color}{0.162 (0.00\%)} \\
BM2820-1 & 0.238 (50.35\%) & 0.193 (22.10\%) & 0.159 (0.37\%) & \textcolor{res2color}{0.159 (0.34\%)}  & 0.159 (0.52\%) & \textcolor{res1color}{0.158 (0.00\%)} \\
BM2820-2 & 0.201 (60.13\%) & 0.163 (29.80\%) & 0.127 (0.90\%) & 0.127 (0.55\%) & \textcolor{res2color}{0.127 (0.54\%)}  & \textcolor{res1color}{0.126 (0.00\%)} \\
BM2820-3 & 0.227 (47.98\%) & 0.186 (21.13\%) & 0.154 (0.16\%) & \textcolor{res2color}{0.154 (0.15\%)}  & 0.154 (0.31\%) & \textcolor{res1color}{0.153 (0.00\%)} \\
BM4018S118-1 & 0.211 (11.76\%) & 0.208 (10.18\%) & \textcolor{res2color}{0.189 (0.03\%)}  & 0.189 (0.08\%) & 0.189 (0.15\%) & \textcolor{res1color}{0.189 (0.00\%)} \\
BM4018S118-2 & 0.247 (8.79\%) & 0.245 (8.19\%) & 0.227 (0.04\%) & \textcolor{res2color}{0.227 (0.02\%)}  & 0.227 (0.03\%) & \textcolor{res1color}{0.227 (0.00\%)} \\
BM4018S118-3 & 0.207 (13.69\%) & 0.205 (12.41\%) & \textcolor{res2color}{0.182 (0.02\%)}  & 0.182 (0.08\%) & 0.183 (0.17\%) & \textcolor{res1color}{0.182 (0.00\%)} \\
GOPRO-1 & 0.201 (36.84\%) & 0.150 (2.17\%) & \textcolor{res2color}{0.147 (0.04\%)}  & 0.147 (0.06\%) & 0.147 (0.30\%) & \textcolor{res1color}{0.147 (0.00\%)} \\
GOPRO-2 & 0.165 (30.52\%) & 0.128 (1.32\%) & \textcolor{res1color}{0.127 (0.00\%)}  & \textcolor{res2color}{0.127 (0.02\%)}  & 0.127 (0.25\%) & 0.127 (0.13\%)\\
GOPRO-3 & 0.235 (40.41\%) & 0.171 (2.17\%) & \textcolor{res2color}{0.167 (0.09\%)}  & 0.168 (0.41\%) & 0.169 (1.02\%) & \textcolor{res1color}{0.167 (0.00\%)} \\
EUROC & 0.137 (4.64\%) & 0.133 (1.21\%) & \textcolor{res2color}{0.131 (0.19\%)}  & 0.131 (0.25\%) & 0.132 (0.31\%) & \textcolor{res1color}{0.131 (0.00\%)} \\
\hline
\end{tabular}
}
\caption{Mean reprojection error for evaluated camera models (in pixels). Best and second-best results are shown in green and orange, respectively. The table also shows overhead in \% compared to the model with the smallest reprojection error in the sequence. The results show that the proposed model, despite having only six parameters, has less than 1\% greater mean reprojection error than the best performing model with eight parameters.}
\label{tab:projection}
\end{table*}

\section{Double Sphere Camera Model}
\label{sec:double_sphere}

We propose the Double Sphere (\textbf{DS}) camera model that better fits cameras with fisheye lenses, has a closed-form inverse, and does not require computationally expensive trigonometric operations. In the proposed DS model a point is consecutively projected onto two unit spheres with centers shifted by $\xi$. Then, the point is projected onto the image plane using the pinhole model shifted by $\frac{\alpha}{1-\alpha}$ (Figure \ref{fig:double_sphere}). This model has six parameters $\mathbf{i} = \left[f_x, f_y, c_x, c_y, \xi, \alpha \right]^T$ and a projection function defined as follows:

\begin{align}
    \pi(\mathbf{x}, \mathbf{i}) &=
    \begin{bmatrix}
    f_x{\frac{x}{\alpha d_2 + (1-\alpha) (\xi d_1 + z)}}\\
    f_y{\frac{y}{\alpha d_2 + (1-\alpha) (\xi d_1 + z)}}\\
    \end{bmatrix}
    +
    \begin{bmatrix}
    c_x \\ 
    c_y\\
    \end{bmatrix}, \\
    d_1 &= \sqrt{x^2 + y^2 + z^2}, \\
    d_2 &= \sqrt{x^2 + y^2 + (\xi  d_1 + z)^2}.
\end{align}
A set of 3D points that results in valid projection is expressed as follows:
\begin{align}
\Omega &= \{\mathbf{x} \in \mathbb{R}^3 ~|~ z > -w_2 d_1 \} \\
w_2 &= \frac{w_1+\xi}{\sqrt{2w_1\xi + \xi^2 + 1}} \\
w_1 &= \begin{cases} \frac{\alpha}{1-\alpha}, & \mbox{if } \alpha \le 0.5 \\  \frac{1-\alpha}{\alpha} & \mbox{if } \alpha > 0.5 \end{cases}
\end{align}

The unprojection function is computed as follows:
\begin{align}
    \pi^{-1}(\mathbf{u}, \mathbf{i}) &= 
    \frac{m_z \xi + \sqrt{m_z^2 + (1 - \xi^2) r^2}}{m_z^2 + r^2}
    \begin{bmatrix}
    m_x \\ 
    m_y \\
    m_z \\
    \end{bmatrix}-\begin{bmatrix}
    0 \\ 
    0 \\
    \xi \\
    \end{bmatrix}, \\
    m_x &= \frac{u - c_x}{f_x}, \\
    m_y &= \frac{v - c_y}{f_y}, \\
    r^2 &= m_x^2 + m_y^2, \\
    m_z &= \frac{1 - \alpha^2  r^2}{\alpha  \sqrt{1 - (2 \alpha - 1)  r^2} + 1 - \alpha},
\end{align}
where the following holds.
\begin{align}
\Theta &= \begin{cases} 
\mathbb{R}^2 & \mbox{if } \alpha \le 0.5 \\
\{ \mathbf{u} \in \mathbb{R}^2 ~|~ r^2 \le \frac{1}{2\alpha-1} \}  & \mbox{if } \alpha > 0.5
\end{cases}
\end{align}

\section{Calibration}
\label{sec:calibration}

To estimate the camera parameters of each model we use a grid of AprilTag markers \cite{olson2011tags} (Figure \ref{fig:double_sphere}) that can be detected automatically in the images. For each image $n$ in the calibration sequence, the detection gives us the 2D position $\mathbf{u}_{nk}$ of the projection of corner $k$ onto the image plane and the associated 3D location $\mathbf{x}_k$ of the corner. After initial marker detection we use local subpixel refinement for each corner to achieve better calibration accuracy.

We formulate the optimization function that depends on the state $\mathbf{s}= \left[\mathbf{i}, {\bf T}_{ca_1}, ..., {\bf T}_{ca_N}\right]$ as follows:
\begin{align}
    E(s) &= \sum_{n=1}^N{ \sum_{k \in K}{ \rho \left( (\pi({\bf T}_{ca_n} \mathbf{x}_k, \mathbf{i}) - \mathbf{u}_{nk} )^2 \right), }}
\end{align}
where $\mathbf{i}$ is the vector of intrinsic parameters, $\pi$ is the projection function, ${\bf T}_{ca_n} \in SE(3)$ is the transformation from the coordinate frame of the calibration grid to the camera coordinate frame for image $n$. $K$ is a set of detected corner points for the image $n$ and $\rho$ is the robust Huber norm.

We parameterize the updates to the state with vector $\Delta \mathbf{s} = \left[\Delta \mathbf{i}, \Delta \mathbf{t}_0, ... , \mathbf{t}_N\right]^T$ as follows:
\begin{align}
  \mathbf{s} \oplus \Delta \mathbf{s} &=
  \begin{bmatrix}
   \mathbf{i} + \Delta \mathbf{i} \\
    {\bf T}_{ca_1} \exp(\Delta \mathbf{t}_1) \\
     ...\\
    {\bf T}_{ca_N}  \exp(\Delta \mathbf{t}_N) \\
  \end{bmatrix}
\end{align}

Given the current state $\mathbf{s}_l$ we can rewrite the optimization function as:
\begin{align}
    E(\mathbf{s}_l \oplus \Delta \mathbf{s}) &= \mathbf{r}(\mathbf{s}_l \oplus \Delta \mathbf{s})^T {\bf W} \mathbf{r}(\mathbf{s}_l \oplus \Delta \mathbf{s}),
\end{align}
and use the Gauss-Newton algorithm to compute the update for the current iteration as follows:
\begin{align}
  \Delta \mathbf{s} &= ({\bf J}_l^T {\bf W} {\bf J}_l)^{-1} {\bf J}_l^T {\bf W} \mathbf{r}_l,
\end{align}
where $\mathbf{r}_l$ is a stacked vector of residuals evaluated at $\mathbf{s}_l$, ${\bf J}_l$ is the Jacobian of residuals with respect to $\Delta \mathbf{s}$, and ${\bf W}$ is the weighting matrix corresponding to the Huber norm. 
With that, we update the current estimate of the state
\begin{align}
  \mathbf{s}_{l+1} &= \mathbf{s}_l \oplus \Delta \mathbf{s},
\end{align}
and iterate until convergence.

Since the optimization function is non-convex, good initialization of the intrinsic parameters $\mathbf{i}$ and camera poses  ${\bf T}_{ca}$ is important for optimization to converge. We initialize the intrinsic parameters with using the previously proposed method \cite{heng2015self} (with $\beta=1$ for EUCM and $\xi=0$ for DS) and find initial poses using the UPnP algorithm \cite{kneip2014upnp}.

\section{Evaluation}
\label{sec:evaluation}

We evaluate the presented camera models using a dataset with 16 sequences. This dataset contains calibration sequences captured with five different lenses (three sequences for each lens) and one calibration sequence from the EuRoC dataset \cite{burri2016_eurocmavdataset}. The lenses used to collect the sequences are shown in Figure \ref{fig:lenses}. To ensure fair comparison, we first detect the calibration corners from all sequences and perform the optimization described in Section \ref{sec:calibration} using the same data for all models.

\begin{table}[t]
\resizebox{\linewidth}{!}{%
\begin{tabular}{ | p{2.2cm} | p{1cm} | p{1.1cm} | p{1cm} | p{1.1cm} | p{1.1cm}  | p{1.1cm} |}
\hline
\textbf{\makecell{Expressions \\ Computed}} &
\textbf{\makecell{UCM}} &
\textbf{\makecell{FOV}}  &
\textbf{\makecell{DS}} & 
\textbf{\makecell{EUCM}} &
\textbf{\makecell{KB 6}} & 
\textbf{\makecell{KB 8}}  \\
\hline
$\pi(\mathbf{x}, \mathbf{i})$ & 33.842 & 419.339 & 55.020 & 32.965 & 288.003 & 305.841 \\
$\pi(\mathbf{x}, \mathbf{i}), ~{\bf J}_{\mathbf{x}}, ~{\bf J}_{\mathbf{i}}$ & 34.555 & 433.956 & 55.673 & 33.534 & 293.625 & 310.399 \\
$\pi^{-1}(\mathbf{u}, \mathbf{i})$ & 71.945 & 430.109 & 107.054 & 92.735 & 561.174 & 638.150 \\
$\pi^{-1}(\mathbf{u}, \mathbf{i}), ~{\bf J}_{\mathbf{u}}, ~{\bf J}_{\mathbf{i}}$ & 71.079 & 891.556 & 181.119 & 95.883 & 537.291 & 613.287 \\
\hline
\end{tabular}
}
\caption{Timing for 10000 operations in microseconds measured on Intel Xeon E5-1620. ${\bf J}$ denotes the Jacobian of the function. The results demonstrate that with similar accuracy, our model shows around five times faster computation time for the projection function than the KB 8 model.}
\label{tab:timing_cam}
\end{table}

\begin{figure*}
    \centering
    \begin{subfigure}[b]{0.32\textwidth}
        \includegraphics[width=\textwidth]{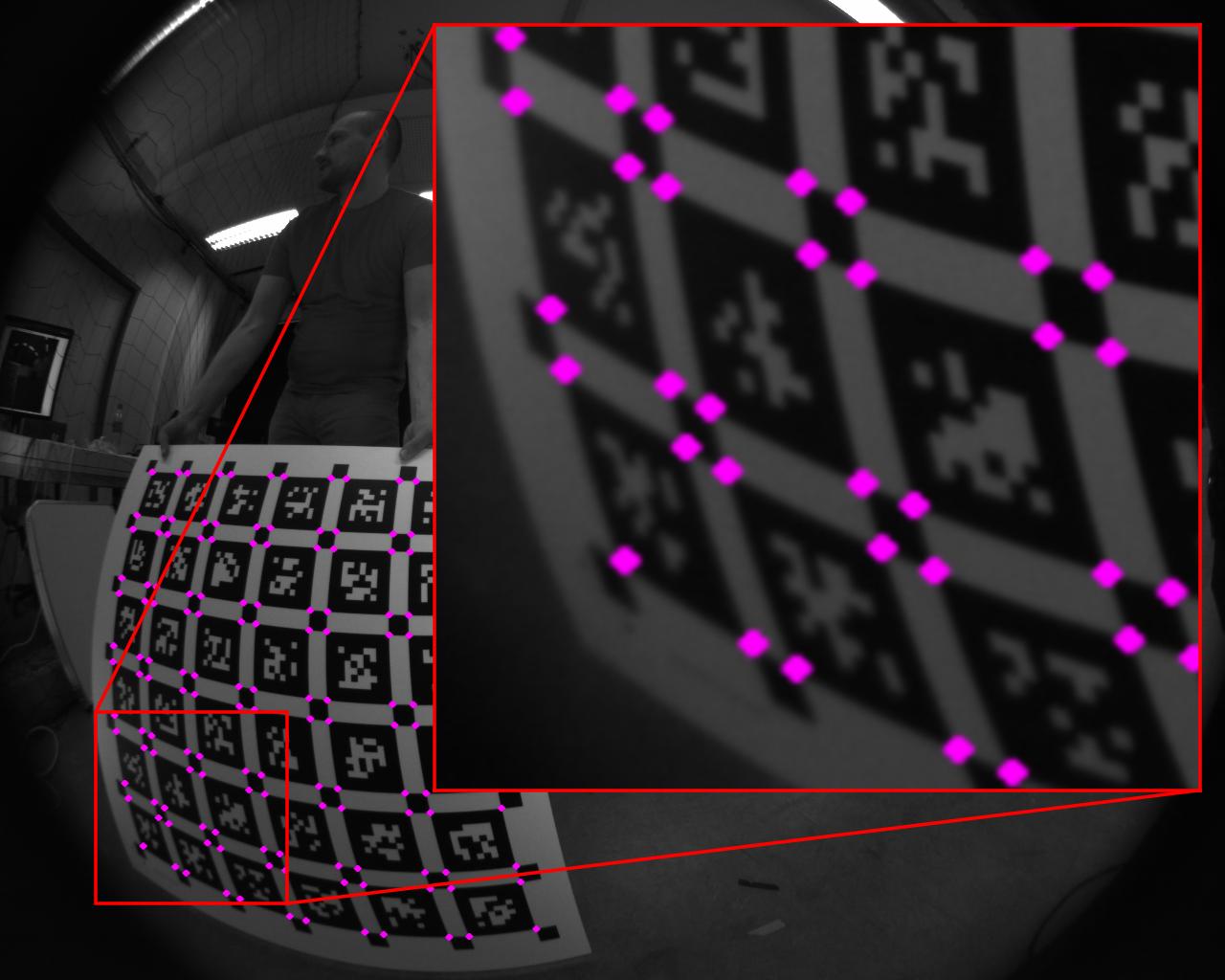}
        \caption{UCM}
    \end{subfigure}
    \begin{subfigure}[b]{0.32\textwidth}
        \includegraphics[width=\textwidth]{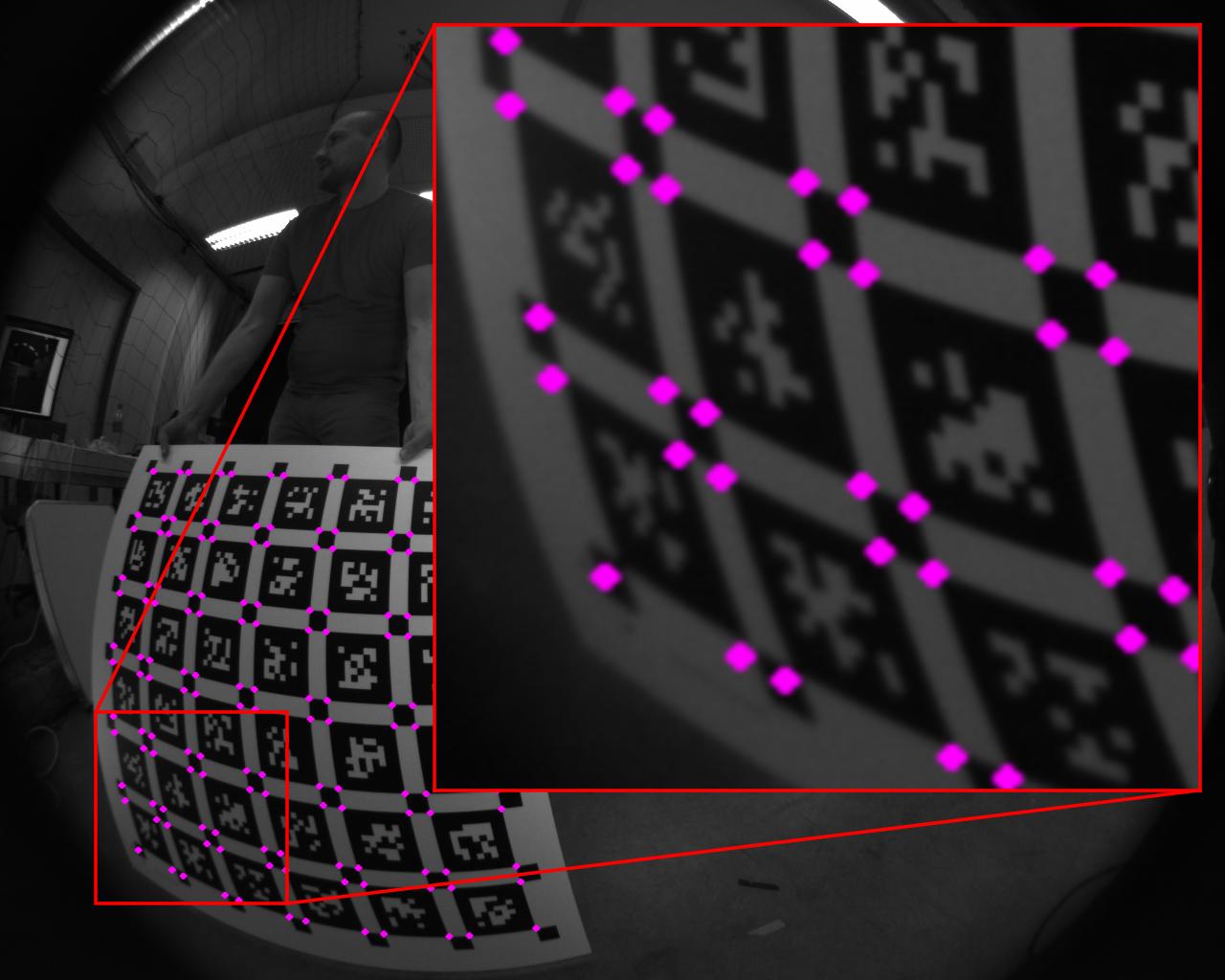}
        \caption{FOV}
    \end{subfigure}
    \begin{subfigure}[b]{0.32\textwidth}
        \includegraphics[width=\textwidth]{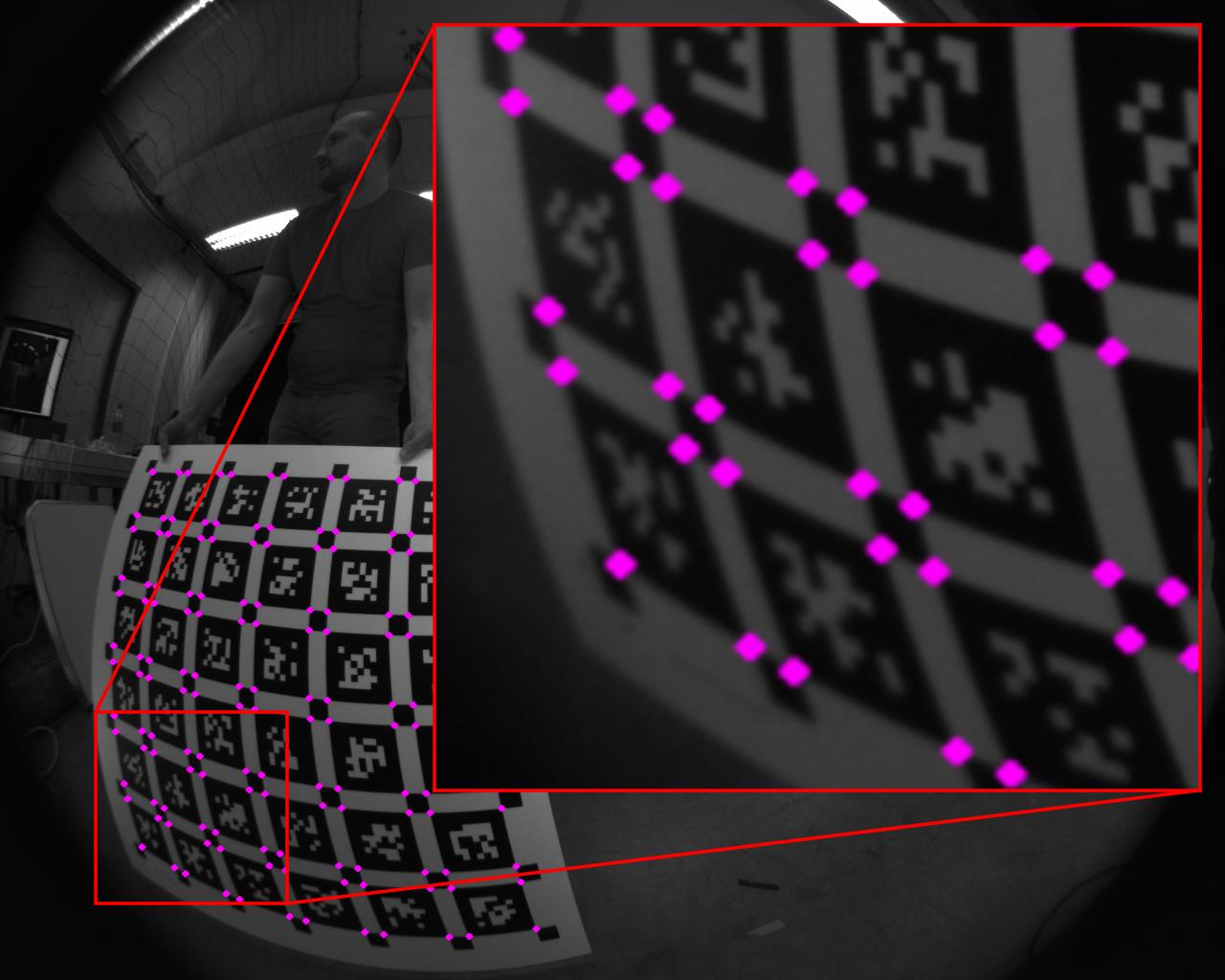}
        \caption{DS}
    \end{subfigure}
    \begin{subfigure}[b]{0.32\textwidth}
        \includegraphics[width=\textwidth]{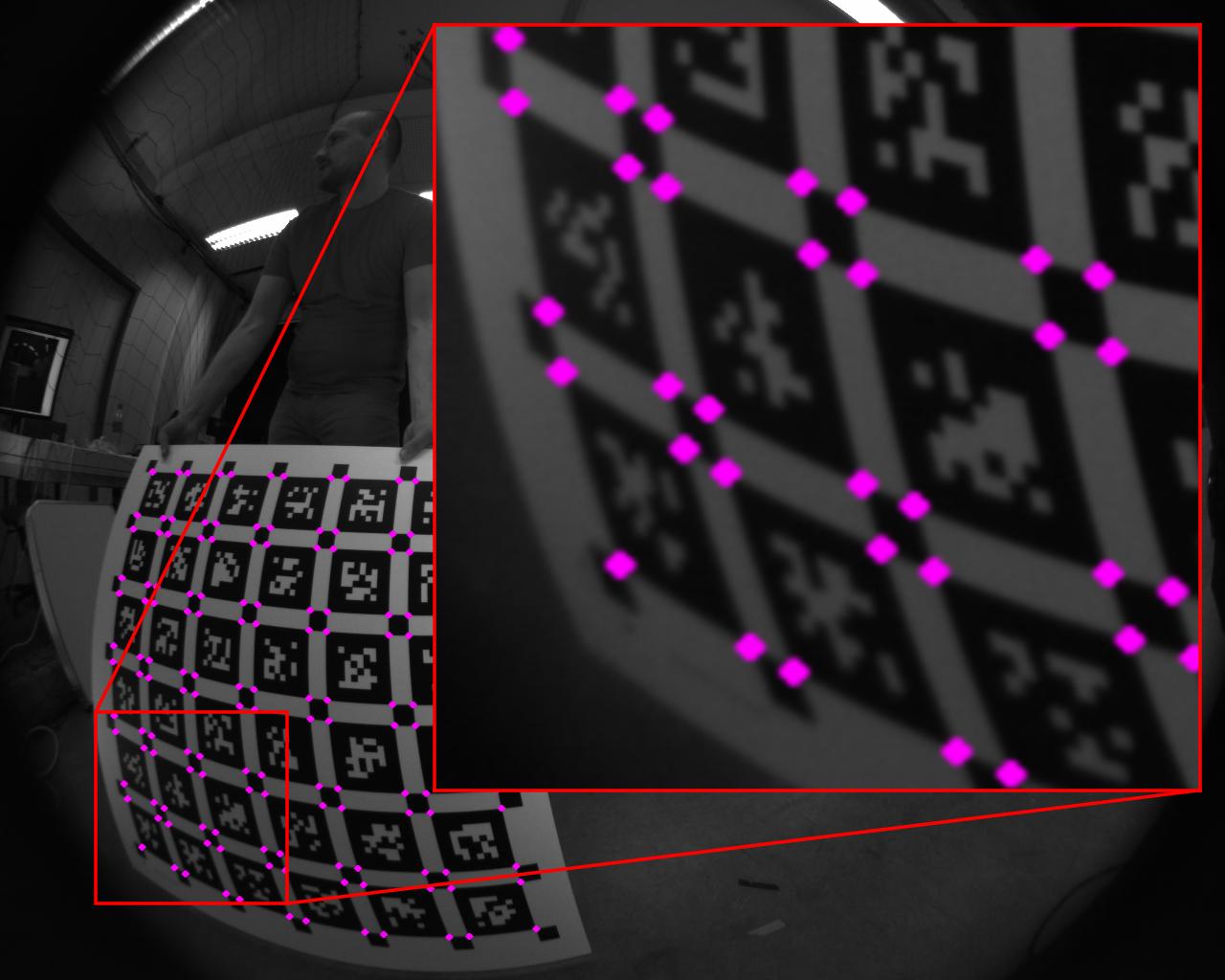}
        \caption{EUCM}
    \end{subfigure}
    \begin{subfigure}[b]{0.32\textwidth}
        \includegraphics[width=\textwidth]{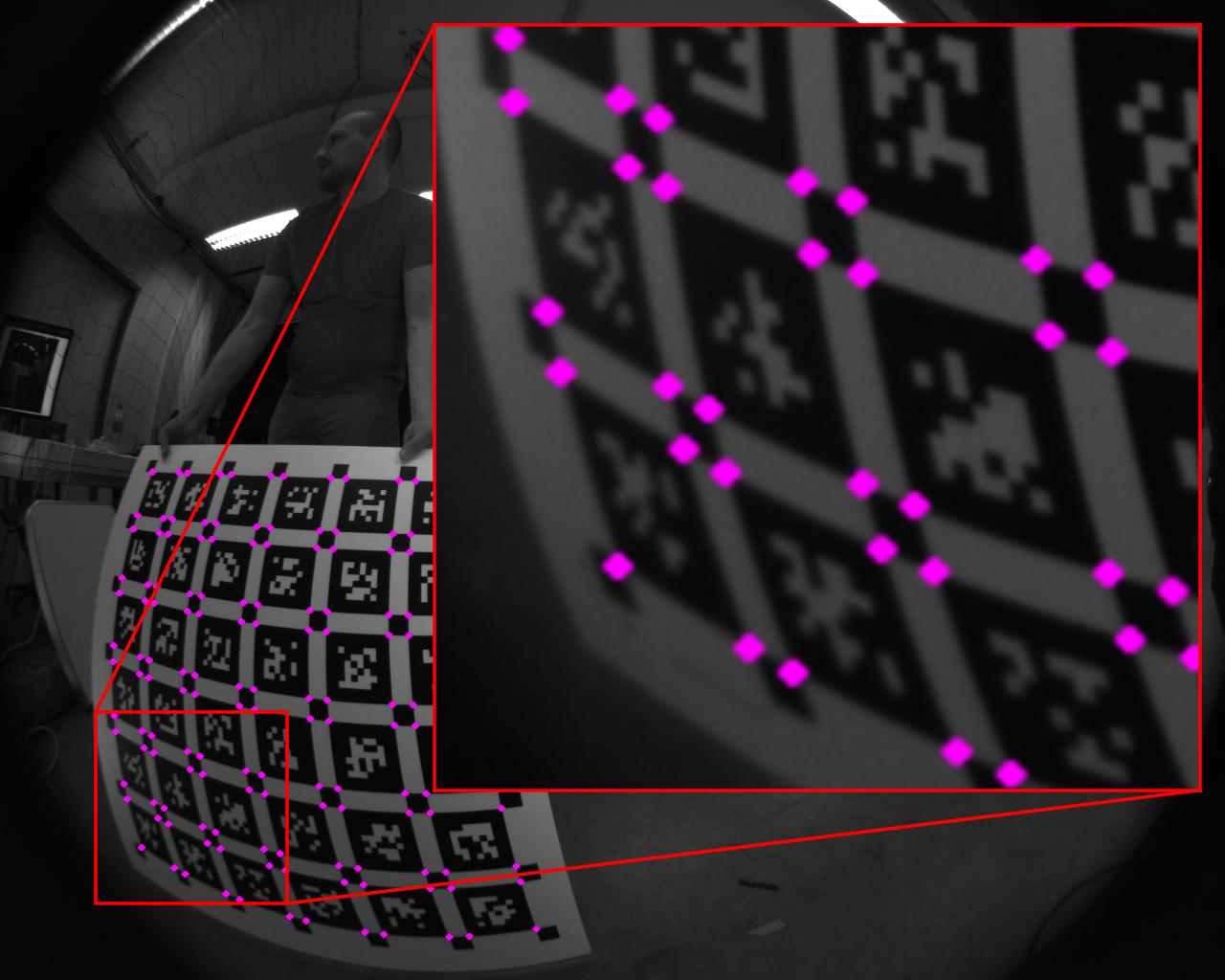}
        \caption{KB 6}
    \end{subfigure}
    \begin{subfigure}[b]{0.32\textwidth}
        \includegraphics[width=\textwidth]{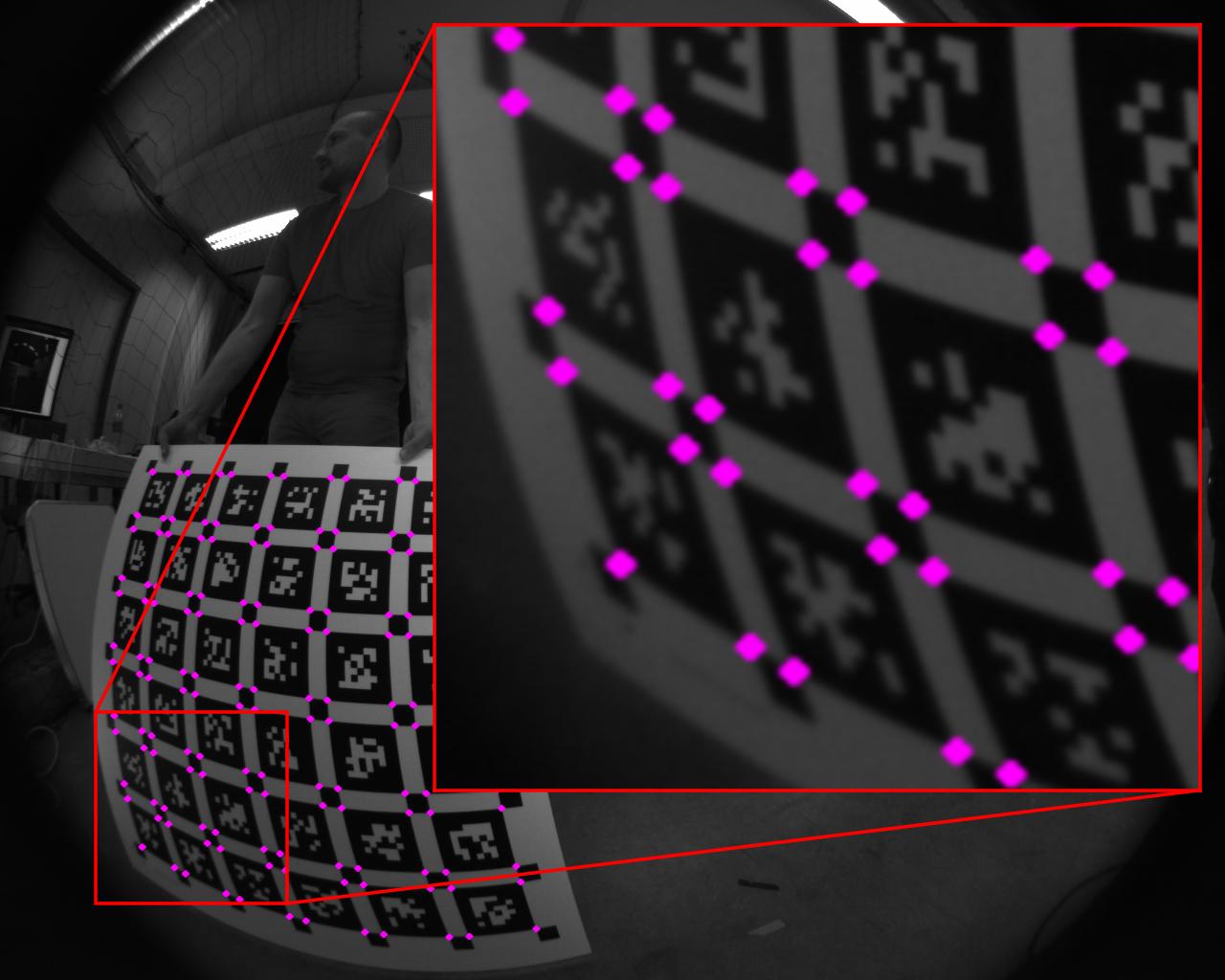}
        \caption{KB 8}
    \end{subfigure}

    \caption{Corners of the calibration pattern (purple) projected onto the image after optimizing camera poses and intrinsic parameters for different camera models. The DS, EUCM and KB 8 models show high reprojection accuracy, while the UCM and KB 6 models have slightly shifted corner positions at the bottom-left corner of the calibration pattern. For the FOV model, displacement of the bottom-left corner is clearly visible, which indicates this model does not well fit the lens.}\label{fig:qualitative}
\end{figure*}

\paragraph{Reprojection error} which indicates how well a model can represent the projection function of the actual lens, is one of the most important metrics for a camera model.
Table \ref{tab:projection} shows the mean reprojection error after optimizing for poses and intrinsic parameters computed for all datasets using different camera models. The best and second-best results for each sequence are shown in green and orange, respectively. For all entries, we also provide overhead computed as $\frac{c-b}{b} \times 100\%$, where $b$ is the smallest reprojection error in the sequence and $c$ is the reprojection error of the current model.

With most of the sequences, the KB model with eight parameters shows the best result, and the proposed model (DS) is the second best. Despite the fact the KB model has eight intrinsic parameters compared to six in the proposed DS model, the reprojection error overhead is less than 1\% for all sequences. The EUCM demonstrates slightly greater reprojection error than that of the DS model and smaller reprojection error than the KB model with six parameters. The UCM and FOV models show greater reprojection errors among all tested models.

\begin{table*}[t]
\resizebox{\textwidth}{!}{%
\begin{tabular}{ | p{2.1cm} | p{3.5cm} | p{3.7cm} | p{3.5cm} | p{3.5cm} | p{3.5cm} | p{4.0cm}  | p{4.2cm} |}
\hline
\textbf{\makecell{Lens}} &
\textbf{\makecell{UCM \\ $\left[f_x, f_y, c_x, c_y, \alpha \right]^T$  }} &
\textbf{\makecell{UCM \\ $\left[\gamma_x, \gamma_y, c_x, c_y, \xi \right]^T$ }} &
\textbf{\makecell{FOV \\ $\left[f_x, f_y, c_x, c_y, w\right]^T$}}  &
\textbf{\makecell{DS \\ $\left[f_x, f_y, c_x, c_y, \xi, \alpha \right]^T$}} & 
\textbf{\makecell{EUCM \\ $\left[f_x, f_y, c_x, c_y, \alpha, \beta \right]^T$}} &
\textbf{\makecell{KB 6 \\ $\left[f_x, f_y, c_x, c_y, k_1, k_2 \right]^T$}} & 
\textbf{\makecell{KB 8 \\ $\left[f_x, f_y, c_x, c_y, k_1, k_2, k_3, k_4 \right]^T$}}  \\
\hline
\rule{0pt}{2cm}

BF2M2020S23 & $ \begin{pmatrix} 377.60 \\ 377.48 \\ 638.74 \\ 514.00 \\ 0.64 \end{pmatrix} \pm \begin{pmatrix} 0.20\% \\ 0.23\% \\ 0.05\% \\ 0.04\% \\ 0.16\% \end{pmatrix} $  & $ \begin{pmatrix} 1041.97 \\ 1041.63 \\ 638.74 \\ 514.00 \\ 1.76 \end{pmatrix} \pm \begin{pmatrix} 0.48\% \\ 0.50\% \\ 0.05\% \\ 0.04\% \\ 0.44\% \end{pmatrix} $  & $ \begin{pmatrix} 352.58 \\ 352.72 \\ 638.23 \\ 513.08 \\ 0.93 \end{pmatrix} \pm \begin{pmatrix} 0.17\% \\ 0.16\% \\ 0.08\% \\ 0.20\% \\ 0.02\% \end{pmatrix} $  & $ \begin{pmatrix} 313.21 \\ 313.21 \\ 638.66 \\ 514.39 \\ -0.18 \\ 0.59 \end{pmatrix} \pm \begin{pmatrix} 0.11\% \\ 0.11\% \\ 0.01\% \\ 0.03\% \\ 0.68\% \\ 0.05\% \end{pmatrix} $  & $ \begin{pmatrix} 380.95 \\ 380.94 \\ 638.66 \\ 514.37 \\ 0.63 \\ 1.04 \end{pmatrix} \pm \begin{pmatrix} 0.04\% \\ 0.04\% \\ 0.01\% \\ 0.03\% \\ 0.03\% \\ 0.06\% \end{pmatrix} $  & $ \begin{pmatrix} 380.14 \\ 380.10 \\ 638.65 \\ 514.30 \\ 0.01 \\ -0.01 \end{pmatrix} \pm \begin{pmatrix} 0.01\% \\ 0.00\% \\ 0.01\% \\ 0.02\% \\ 4.04\% \\ 3.51\% \end{pmatrix} $  & $ \begin{pmatrix} 380.99 \\ 380.98 \\ 638.66 \\ 514.38 \\ 0.01 \\ -0.00 \\ 0.00 \\ -0.00 \end{pmatrix} \pm \begin{pmatrix} 0.02\% \\ 0.03\% \\ 0.01\% \\ 0.03\% \\ 6.35\% \\ 22.04\% \\ 511.08\% \\ 15.18\% \end{pmatrix} $ \\ \rule{0pt}{2cm}

BM2820 & $ \begin{pmatrix} 528.31 \\ 528.46 \\ 624.08 \\ 512.58 \\ 0.64 \end{pmatrix} \pm \begin{pmatrix} 0.20\% \\ 0.19\% \\ 0.05\% \\ 0.03\% \\ 0.33\% \end{pmatrix} $  & $ \begin{pmatrix} 1470.51 \\ 1470.93 \\ 624.08 \\ 512.58 \\ 1.78 \end{pmatrix} \pm \begin{pmatrix} 0.78\% \\ 0.77\% \\ 0.05\% \\ 0.03\% \\ 0.92\% \end{pmatrix} $  & $ \begin{pmatrix} 491.60 \\ 491.71 \\ 624.20 \\ 512.68 \\ 0.92 \end{pmatrix} \pm \begin{pmatrix} 0.14\% \\ 0.14\% \\ 0.05\% \\ 0.04\% \\ 0.14\% \end{pmatrix} $  & $ \begin{pmatrix} 386.17 \\ 386.23 \\ 624.29 \\ 512.49 \\ -0.27 \\ 0.55 \end{pmatrix} \pm \begin{pmatrix} 0.20\% \\ 0.21\% \\ 0.04\% \\ 0.02\% \\ 0.38\% \\ 0.12\% \end{pmatrix} $  & $ \begin{pmatrix} 530.18 \\ 530.27 \\ 624.28 \\ 512.49 \\ 0.57 \\ 1.17 \end{pmatrix} \pm \begin{pmatrix} 0.08\% \\ 0.09\% \\ 0.04\% \\ 0.02\% \\ 0.14\% \\ 0.06\% \end{pmatrix} $  & $ \begin{pmatrix} 530.09 \\ 530.18 \\ 624.28 \\ 512.49 \\ -0.00 \\ 0.01 \end{pmatrix} \pm \begin{pmatrix} 0.09\% \\ 0.09\% \\ 0.04\% \\ 0.02\% \\ 140.31\% \\ 0.26\% \end{pmatrix} $  & $ \begin{pmatrix} 530.35 \\ 530.44 \\ 624.29 \\ 512.48 \\ -0.01 \\ 0.02 \\ -0.02 \\ 0.01 \end{pmatrix} \pm \begin{pmatrix} 0.08\% \\ 0.09\% \\ 0.04\% \\ 0.02\% \\ 15.18\% \\ 14.05\% \\ 20.23\% \\ 21.39\% \end{pmatrix} $ \\ \rule{0pt}{2cm}

BF5M13720 & $ \begin{pmatrix} 258.53 \\ 258.45 \\ 637.53 \\ 511.89 \\ 0.65 \end{pmatrix} \pm \begin{pmatrix} 0.07\% \\ 0.05\% \\ 0.01\% \\ 0.06\% \\ 0.03\% \end{pmatrix} $  & $ \begin{pmatrix} 741.24 \\ 741.03 \\ 637.53 \\ 511.89 \\ 1.87 \end{pmatrix} \pm \begin{pmatrix} 0.10\% \\ 0.09\% \\ 0.01\% \\ 0.06\% \\ 0.07\% \end{pmatrix} $  & $ \begin{pmatrix} 242.16 \\ 242.18 \\ 637.51 \\ 512.21 \\ 0.95 \end{pmatrix} \pm \begin{pmatrix} 0.09\% \\ 0.11\% \\ 0.03\% \\ 0.05\% \\ 0.15\% \end{pmatrix} $  & $ \begin{pmatrix} 208.36 \\ 208.35 \\ 637.45 \\ 512.18 \\ -0.20 \\ 0.59 \end{pmatrix} \pm \begin{pmatrix} 0.14\% \\ 0.14\% \\ 0.01\% \\ 0.02\% \\ 0.41\% \\ 0.06\% \end{pmatrix} $  & $ \begin{pmatrix} 260.67 \\ 260.66 \\ 637.45 \\ 512.17 \\ 0.64 \\ 1.06 \end{pmatrix} \pm \begin{pmatrix} 0.07\% \\ 0.07\% \\ 0.01\% \\ 0.02\% \\ 0.09\% \\ 0.11\% \end{pmatrix} $  & $ \begin{pmatrix} 260.28 \\ 260.27 \\ 637.45 \\ 512.15 \\ 0.00 \\ -0.00 \end{pmatrix} \pm \begin{pmatrix} 0.06\% \\ 0.06\% \\ 0.01\% \\ 0.02\% \\ 23.96\% \\ 3.31\% \end{pmatrix} $  & $ \begin{pmatrix} 260.87 \\ 260.86 \\ 637.45 \\ 512.19 \\ -0.01 \\ -0.00 \\ -0.00 \\ -0.00 \end{pmatrix} \pm \begin{pmatrix} 0.10\% \\ 0.10\% \\ 0.01\% \\ 0.02\% \\ 33.44\% \\ 584.87\% \\ 741.92\% \\ 65.29\% \end{pmatrix} $ \\ \rule{0pt}{2cm}

GOPRO & $ \begin{pmatrix} 499.67 \\ 499.78 \\ 620.72 \\ 513.74 \\ 0.68 \end{pmatrix} \pm \begin{pmatrix} 0.03\% \\ 0.04\% \\ 0.07\% \\ 0.16\% \\ 0.13\% \end{pmatrix} $  & $ \begin{pmatrix} 1546.22 \\ 1546.54 \\ 620.72 \\ 513.74 \\ 2.09 \end{pmatrix} \pm \begin{pmatrix} 0.30\% \\ 0.31\% \\ 0.07\% \\ 0.16\% \\ 0.41\% \end{pmatrix} $  & $ \begin{pmatrix} 462.90 \\ 462.94 \\ 621.07 \\ 513.36 \\ 0.95 \end{pmatrix} \pm \begin{pmatrix} 0.02\% \\ 0.03\% \\ 0.06\% \\ 0.10\% \\ 0.00\% \end{pmatrix} $  & $ \begin{pmatrix} 364.84 \\ 364.86 \\ 621.12 \\ 513.27 \\ -0.27 \\ 0.57 \end{pmatrix} \pm \begin{pmatrix} 0.09\% \\ 0.08\% \\ 0.06\% \\ 0.10\% \\ 0.26\% \\ 0.06\% \end{pmatrix} $  & $ \begin{pmatrix} 501.02 \\ 501.06 \\ 621.12 \\ 513.26 \\ 0.60 \\ 1.17 \end{pmatrix} \pm \begin{pmatrix} 0.03\% \\ 0.04\% \\ 0.06\% \\ 0.10\% \\ 0.22\% \\ 0.26\% \end{pmatrix} $  & $ \begin{pmatrix} 500.92 \\ 500.96 \\ 621.10 \\ 513.26 \\ -0.02 \\ 0.00 \end{pmatrix} \pm \begin{pmatrix} 0.03\% \\ 0.04\% \\ 0.06\% \\ 0.10\% \\ 1.65\% \\ 6.96\% \end{pmatrix} $  & $ \begin{pmatrix} 501.13 \\ 501.17 \\ 621.12 \\ 513.27 \\ -0.02 \\ 0.01 \\ -0.00 \\ 0.00 \end{pmatrix} \pm \begin{pmatrix} 0.04\% \\ 0.05\% \\ 0.06\% \\ 0.09\% \\ 6.47\% \\ 41.92\% \\ 178.38\% \\ 12621.85\% \end{pmatrix} $ \\ \rule{0pt}{2cm}

BM4018S118 & $ \begin{pmatrix} 735.84 \\ 736.03 \\ 635.44 \\ 521.89 \\ 0.62 \end{pmatrix} \pm \begin{pmatrix} 0.09\% \\ 0.08\% \\ 0.06\% \\ 0.02\% \\ 0.14\% \end{pmatrix} $  & $ \begin{pmatrix} 1933.84 \\ 1934.35 \\ 635.44 \\ 521.89 \\ 1.63 \end{pmatrix} \pm \begin{pmatrix} 0.31\% \\ 0.29\% \\ 0.06\% \\ 0.02\% \\ 0.37\% \end{pmatrix} $  & $ \begin{pmatrix} 686.06 \\ 686.22 \\ 635.45 \\ 521.92 \\ 0.90 \end{pmatrix} \pm \begin{pmatrix} 0.10\% \\ 0.09\% \\ 0.06\% \\ 0.01\% \\ 0.09\% \end{pmatrix} $  & $ \begin{pmatrix} 565.58 \\ 565.68 \\ 635.52 \\ 521.81 \\ -0.23 \\ 0.55 \end{pmatrix} \pm \begin{pmatrix} 0.70\% \\ 0.71\% \\ 0.06\% \\ 0.02\% \\ 2.33\% \\ 0.32\% \end{pmatrix} $  & $ \begin{pmatrix} 736.95 \\ 737.08 \\ 635.52 \\ 521.81 \\ 0.57 \\ 1.11 \end{pmatrix} \pm \begin{pmatrix} 0.11\% \\ 0.10\% \\ 0.06\% \\ 0.02\% \\ 0.60\% \\ 0.75\% \end{pmatrix} $  & $ \begin{pmatrix} 736.92 \\ 737.05 \\ 635.52 \\ 521.81 \\ 0.02 \\ 0.00 \end{pmatrix} \pm \begin{pmatrix} 0.11\% \\ 0.10\% \\ 0.06\% \\ 0.02\% \\ 4.26\% \\ 15.24\% \end{pmatrix} $  & $ \begin{pmatrix} 737.02 \\ 737.15 \\ 635.52 \\ 521.81 \\ 0.02 \\ 0.01 \\ -0.00 \\ 0.00 \end{pmatrix} \pm \begin{pmatrix} 0.11\% \\ 0.10\% \\ 0.06\% \\ 0.02\% \\ 5.20\% \\ 21.76\% \\ 78.35\% \\ 191.37\% \end{pmatrix} $ \\

\hline
\end{tabular}
}
\caption{Mean and standard deviation (in \%) of intrinsic parameters computed on three different sequences for each lens. The results suggest that our formulation of  the UCM (first column, compare Eq. \ref{eq:ucm_ours}) has smaller standard deviation compared to standard formulation \cite{mei} (second column), where changes to $\gamma$ and $\xi$ have significant effect on each other.
}
\label{tab:intrinsics}
\end{table*}

\paragraph{Computation time} is another important aspect of a camera model because projection and unprojection functions are evaluated thousands of times in each iteration of vision-based motion estimation. Moreover, for optimization algorithms we must compute the Jacobians of these functions relative to the points and intrinsic parameters; thus, the computation time of these operations should also be considered.

Table \ref{tab:timing_cam} summarizes the computation times of those operations for the presented models. For each camera model, we measure the time of 10000 operations using the Google Benchmark\footnote{\url{https://github.com/google/benchmark}} library on an Intel Xeon E5-1620 CPU. To compile the benchmarks, we use GCC 7 with O3 optimization level and execute the code in a single thread. Note that a small time difference between computing only the function and computing the function with Jacobians can be explained by the superscalar architecture of modern CPUs, which parallelizes execution internally. 

The timing results show that the FOV and KB models are much slower than the other models. For example, the KB model with eight parameters is approximately nine times slower than the EUCM and five times slower than the DS model when evaluating the projection function. This is due to the fact that the KB model involves computationally expensive trigonometric operations ($atan2$).

Unprojection in KB models require iterative optimization to solve the polynomial roots, which together with the trigonometric operations, makes it several times slower than the UCM, EUCM and DS models. The FOV model is the slowest relative to unprojection, which is likely due to its multiple trigonometric operations.

\paragraph{Qualitative results} of reprojection quality for the evaluated models are shown in Figure \ref{fig:qualitative}. Here, we project the corners of the calibration pattern after optimizing for pose and intrinsic parameters and visualize them on the corresponding image taken from the BF2M2020S23-3 sequence. The DS EUCM and KB 8 models provide similar results that are difficult to distinguish by the human eye. The UCM and KB 6 model well fit the corners in the middle of the image; however, these models have a small shift close to the edges. Note that imperfections are clearly visible with the FOV model. 

\paragraph{Different formulations of UCM} are evaluated in terms of the numerical stability of the results. 
\label{sec:comparison_ucm}
Table \ref{tab:intrinsics} shows the mean and standard deviation (in \%) of the intrinsic parameters computed on three different sequences for each lens. For the UCM we provide two formulations with the same reprojection error that are formulated with different intrinsic parameters. The results for the standard formulation as defined in the literature \cite{mei} ($\mathbf{i} = \left[\gamma_x, \gamma_y, c_x, c_y, \xi \right]^T$) are presented in the second column and show higher standard deviation than the results of the model parametrized with $\mathbf{i} = \left[f_x, f_y, c_x, c_y, \alpha \right]^T$. This can be explained by the strong coupling between $\gamma_x, \gamma_y$ and $\xi$, which is not the case for the proposed parametrization. Moreover, for this formulation the focal length stays close to the focal length of the other camera models.

\section{Conclusion}

In this paper, we present the novel Double Sphere camera model that is well suited to fisheye cameras. We compare the proposed camera model to other state-of-the-art camera models. In addition, we provide an extensive evaluation of the presented camera models using 16 different calibration sequences and six different lenses. The evaluation results demonstrate that the model based on high-order polynomials (i.e., KB 8) shows the lowest reprojection error but is 5-10 times slower than competing models. Both the proposed DS model and the EUCM show very low reprojection error, with the DS model being slightly more accurate (less than 1\% greater reprojection error compared to KB 8 on all sequences), and the EUCM being slightly faster (nine times faster projection evaluation than KB 8). Moreover, both models have a closed-form inverse and do not require computationally expensive trigonometric operations.

These results demonstrate that models based on spherical projection present a good alternative to models based on high-order polynomials for applications where fast projection, unprojection and a closed-form inverse are required.

\section*{Acknowledgment}
This work was partially supported by the grant ``For3D" by the Bavarian Research Foundation, the grant CR~250/9-2 ``Mapping on Demand" by the German Research Foundation and the ERC Consolidator Grant ``3D Reloaded".

\balance

{\small
\bibliographystyle{ieee}
\bibliography{egbib}
}

\end{document}